\documentclass[conference]{IEEEtran}
\IEEEoverridecommandlockouts
\usepackage{cite}
\usepackage{amsmath,amssymb,amsfonts}
\usepackage{algorithmic}
\usepackage{graphicx}
\usepackage{xcolor}
\usepackage{times}
\usepackage{graphicx}
\usepackage{amsmath}
\usepackage{amssymb}
\usepackage{comment}
\usepackage{hyperref}
\hypersetup{colorlinks=true,urlcolor=blue,citecolor=blue}
\graphicspath{{grf/}}

\begin{document}

\title{WW-Nets: \\
Dual Neural Networks for Object Detection
}

\author{\IEEEauthorblockN{Mohammad K.~Ebrahimpour$^*$, J.~Ben Falandays$^+$, Samuel Spevack$^+$, Ming-Hsuan Yang$^*$, and David C.~Noelle$^{*+}$}
\IEEEauthorblockA{\textit{$^*$Electrical Engineering \& Computer Science, $^+$Cognitive and Information Sciences} \\
\textit{University of California, Merced}}}

\maketitle

\begin{abstract}
We propose a new deep convolutional neural network framework that uses
object location knowledge implicit in network connection weights to
guide \emph{selective attention} in object detection tasks. Our approach
is called What-Where Nets (WW-Nets), and it is inspired by the structure of human visual
pathways. In the brain, vision incorporates two separate streams, one in
the temporal lobe and the other in the parietal lobe, called the
ventral stream and the dorsal stream, respectively. The ventral
pathway from primary visual cortex is dominated by ``what''
information, while the dorsal pathway is dominated by ``where''
information. Inspired by this structure, we have proposed an object
detection framework involving the integration of a ``What
Network'' and a ``Where Network''. The aim of the What Network is
to provide selective attention to the relevant parts of the input
image. The Where Network uses this information to locate and classify
objects of interest. In this paper, we compare this approach to
state-of-the-art algorithms on the PASCAL VOC 2007 and 2012 and COCO object detection challenge datasets. Also, we compare out approach to human ``ground-truth" attention. We report the results of an eye-tracking experiment on human subjects using images from PASCAL VOC 2007,
and we demonstrate interesting relationships between human overt
attention and information processing in our WW-Nets. Finally, we
provide evidence that our proposed method performs favorably in
comparison to other object detection approaches, often by a large
margin. The code and the eye-tracking ground-truth dataset can be found at: \url{https://github.com/mkebrahimpour}. 
\end{abstract}


\section{Introduction}
\label{s:introduction}
In recent years, deep Convolutional Neural Networks (CNNs) have been
shown to be effective at image classification, accurately performing
object recognition even in cases involving a large array of object
classes, given a sufficiently rich dataset of 
images~\cite{alexnet,vgg16,seNet,denseNet}.

\begin{figure*}[t]
  \centering
    \includegraphics[width=0.99\linewidth]{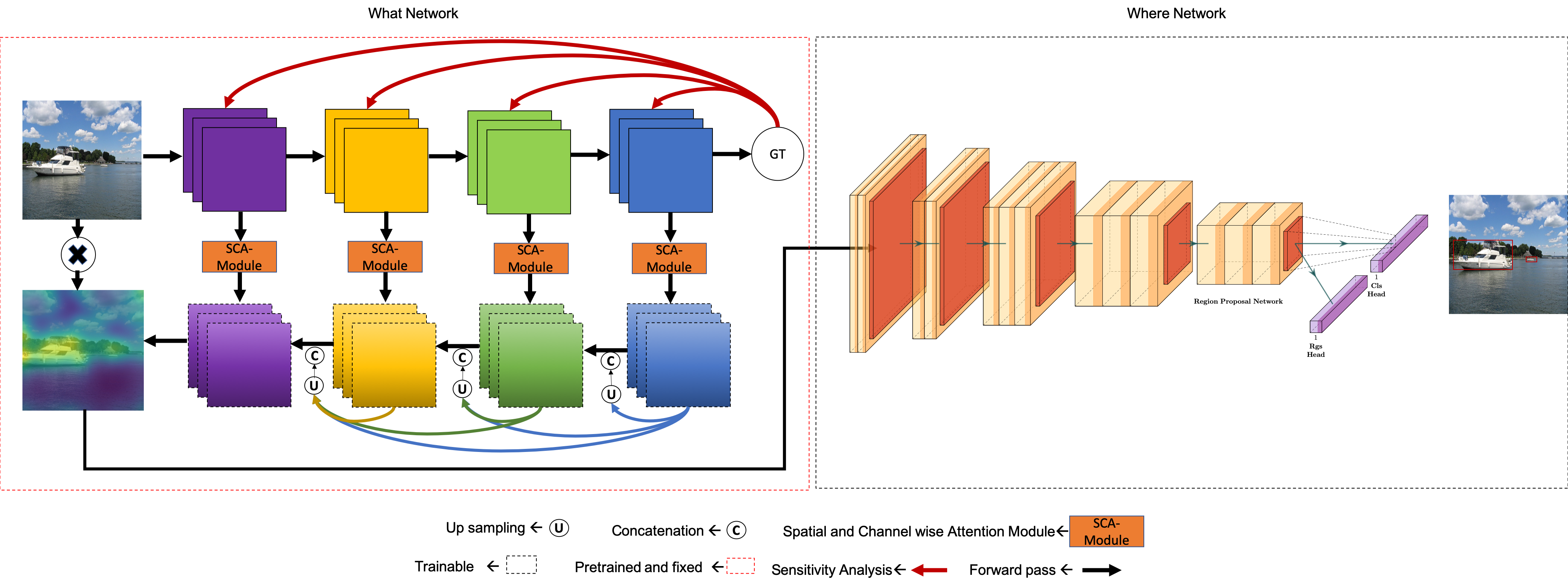}
  \caption{WW-Nets architecture. The What Net guides the selective attention via combining the Spatial and Channel wise Attention in all convolutional layers (SCA). It also leverages from the hidden semantic information in the late layers as well as the hidden location information in the earlier layers by dense connections. Then the filtered image feeds into the Where Net for drawing the bounding boxes out of all objects of interest.} 
  \label{f:DCA}
\end{figure*}

Image classification is only one of the core problems of computer
vision, however. Beyond object
recognition~\cite{vgg16,seNet,denseNet}, there are applications for
such capabilities as semantic segmentation~\cite{semanticSeg1,semanticSeg2,maskrcnn2017}, image captioning~\cite{caption1,caption2,showAttendTell,ebrahimpour_caption}, and object detection~\cite{fastrcnn,fasterrcnn,yolo,yolo9000,ebrahimpour_sensitivity}. The last of these
involves locating and classifying all of the relevant objects in an
image. This is a challenging problem that has received a good deal of
attention~\cite{fasterrcnn,fastrcnn,rcnn,yolo}. Since there is rarely \emph{a priori} information about where objects are located in an
image, most approaches to object detection conduct exhaustive searches over image regions, seeking objects of interest with different sizes and aspect
ratios. For example, region proposal frameworks, like
Faster-RCNN~\cite{fasterrcnn}, need to pass a large number of
candidate image regions through a deep network in order to determine
which parts of the image contain the most information concerning
objects of interest. An alternative approach involves one shot
detectors, like Single Shot Detectors (SSD)~\cite{ssd} and You Only
Look Once (YOLO)~\cite{yolo}. These methods use networks to examine
all parts of the image via a tiling mechanism. For example, YOLO
conducts a search over potential combinations of tiles. The hope of
single shot detection approaches is to find the responsible tile for
the object and then identify the appropriate object location bounding
box around that tile. In a sense, most off-the-shelf object detection
algorithms distribute attention to all parts of the image equally. Allocating equal attention across the scene is not common in humans, however. Our brain has evolved pay selective attention to the vital parts of the visual scene~\cite{selective_attention}. 

Recently,  an object detection framework inspired by the human visual system has been proposed, called Ventral-Dorsal Nets or, simply, VDNets~\cite{VDNet}. 
In VDNets, a sensitivity analysis in a pretrained image classification network (the Ventral Net) is used to guide localization, and this approach shows promise. However, mistakes made by the Ventral Net can be catastrophic, masking out image regions that contain objects of interest, making their detection impossible. Moreover, despite the fact that this approach is accurate and faster than typical region proposal based object detection algorithms, it still cannot process images in real time.

In this paper, we propose a new method for extracting dense information
about object location from pretrained networks, with the goal of
improving selective attention. Specifically, these new What-Where Nets (WW-Nets) make use of channel-wise attention levels, as well as spatially specific attentional information from all receptive fields. By
stacking these different kinds of attention maps, we can potentially
preserve the semantic (object class) information that comes from late
layers in the network while incorporating spatial location information
that is richly represented in early layers. By using these stacked
attention maps~\cite{denseNet}, our selective attention method is substantially different than a simple sensitivity analysis. We have found that our selective attention mechanism can also substantially improve object detection performance. Moreover, we trained a single shot detector on top of our salience map mechanism, which made the WW-Nets suitable for real time applications.

The human visual attention system supports our naturally strong visual
perception capabilities, so we see it as a useful guide for assessing
selective attention behavior. Thus, in addition to measuring the
object detection performance of our proposed system, we investigated
possible relationships between information in WW-Nets and the patterns of attention exhibited by humans observing the same images. We used
the distribution of fixation points produced by human subjects,
measured using eye-tracking technology, as a measure of how the visual
system distributes attention over images. The human visual attention data will be available publicly for future research efforts.
The contribution of this work can be summarized as follows:
\begin{itemize}
\item Our WW-Nets include a sophisticated, learning free, mechanism to obtain information from different levels of the CNN. It assigns a weight to every single spatial location and channel in every convolutional layer. Also, it leverages the information from the most abstract features as well as the hidden location information in the earlier layers by stacking them together.
\item We show that, when identifying objects in benchmark image datasets, this framework provides superior object detection performance over comparison methods, often by a large margin.
\item We report human attention data collected via an eye-tracking study to provide ``ground-truth" information for vision researchers interested in where people look.
\item We provide comparisons of the attention of  WW-Nets to human attention.
\end{itemize}
\section{Related Work}
 \label{s:related-work}
\textbf{Attention-Based Object Detection.}
Attention-based object detection methods depend on a set of training
images with associated class labels but \emph{without} any object
location bounding box annotations~\cite{ebrahimpour_attention,ebrahimpour_sensitivity}. The lack of a need for ground-truth
bounding boxes is a substantial benefit of this approach, since
manually obtaining such information is costly. 

One object detection approach of this kind is the Class Activation Map
(CAM) method~\cite{cam2016}. This approach is grounded in the
observation that the fully connected layers that appear near the
output of typical CNNs largely discard spatial information. 
To compensate for this, the last convolutional layer is scaled up to the
size of the original image, and Global Average Pooling (GAP) is
applied to the result. 
A linear transformation from the Global Average Pooling (GAP) values to class labels is learned. The learned weights
to a given class output are taken as indicating the relative
importance of different filters for identifying objects of that
class.

For a given image, the individual filter activation patterns in
the upscaled convolutional layer are entered into a weighted sum,
using the linear transformation weights for a class of interest. The
result of this sum is a Class Activation Map that reveals image
regions associated with the target class.

The object detection success of the CAM method has been demonstrated,
but it has also inspired alternative approaches. The work of Selvaraju
et al.~\cite{gradcam2016} suggested that Class Activation Maps could
be extracted from standard image classification networks without any
modifications to the network architecture and additional training to
learn filter weights. The proposed Grad-CAM method computes the
gradients of output labels with respect to the last convolutional
layer, and these gradients are aggregated to produce the filter
weights needed for CAM generation. This is an excellent example of
saliency based approaches that interpret trained deep CNNs, with
others also reported in the
literature~\cite{scale-transfer,progressive,W2F}.

As previously noted, attention-based object detection methods benefit
from their lack of dependence on bounding box annotations on training
images. They also tend to be faster than supervised object detection
approaches, producing results by interpreting the internal weights and
activation maps of an image classification CNN. However, these methods
have been found to be less accurate than supervised object detection
techniques.\\
\\
\textbf{Supervised Object Detection.}
Supervised object detection approaches require training data that
include both class labels and tight bounding box annotations for each
object of interest. Explicitly training on ground-truth bounding boxes
tends to make these approaches more accurate than weakly supervised
methods. These approaches tend to be computationally expensive,
however, due to a need to search through the space of image regions,
processing each region with a deep CNN. Tractability is sought by
reducing the number of image regions considered, selecting from the
space of all possible regions in an informed manner. Methods vary in
how the search over regions is constrained.

Some algorithms use a region proposal based framework. A deep
CNN is trained to produce both a classification output and location
bounding box coordinates, given an input image. Object detection is
performed by considering a variety of rectangular regions in the
image, training the CNN class output when an object of interest is in
the input region presented to the network. Importantly, rather than
consider all possible regions, the technique depends on a \emph{region
proposal algorithm} to identify the image regions to be processed by
the CNN. The region proposal method could be either an external
algorithm like Selective Search~\cite{selective-search}, or it could
be an internal component of the network, as done in
Faster-RCNN~\cite{fasterrcnn}. The most efficient object detection
methods of this kind are R-CNN~\cite{rcnn}, Fast-RCNN~\cite{fastrcnn},
Faster-RCNN~\cite{fasterrcnn}, and
Mask-RCNN~\cite{maskrcnn2017}. Systems using this framework tend to be
quite accurate, but they face a number of challenges beyond issues of 
speed. For example, in an effort to propose regions containing objects
of known classes, it is common to base region proposals on information
appearing late in the network, such as the last convolutional
layer. The lack of high resolution spatial information late in the
network makes it difficult to detect small objects using this
approach. There are a number of research projects that aim to address
this issue by combining low level features and high level ones in
various ways~\cite{scale-transfer,feature-pyramid}.

Rather than incorporating a region proposal mechanism, some supervised
methods perform object detection in one feed-forward pass. A
prominent method of this kind is YOLO, as well as its
extensions~\cite{yolo,yolo9000}. In this approach, the image is
divided into tiles, and each tile is annotated with anchor boxes of
various sizes, proposing relevant regions. The resulting information,
along with the image tiles, are processed by a deep network in a
single pass in order to find all objects of interest. While this
technique is less accurate than region proposal approaches like
Faster-RCNN, it is much faster, increasing its utility for online
applications.

It is worth noting that supervised object detection methods can be seen as spreading attention across the full image, examining all possible regions, to some degree. This is computationally costly.

A comparison of these two general approaches to object detection
displays a clear trade-off between accuracy and computational cost
(speed). This gives rise to the question of whether this trade-off can
\vspace{2mm}
be avoided, in some way.\\
\noindent \textbf{Dual Neural Networks as Two Pathways for Object Detection.}
As previously noted, our brain evolved to pay attention to the gist of the scene and ignore the unimportant parts~\cite{selective_attention}. This is one of the reasons why human beings are good at finding objects in images. Neither of the two general frameworks for object detection take advantage of such a mechanism.
Recently, a novel object detection framework called Ventral-Dorsal Networks (VDNets) has been proposed as an
object detection approach inspired by the human visual system. VDNets
are actually composed of two interacting deep networks, reflecting the
two major information processing streams emerging from primary visual
cortex~\cite{VDNet}.  
We have found that VDNets exhibit strong object detection performance, but
they can fail catastrophically if the Ventral Network mistakenly masks
out objects of interest. Moreover, despite an increase in detection speed, this method cannot be used in real time applications due to a bottleneck in the Dorsal Network. 

With the goal of improving Ventral Network performance, this paper
proposes a substantially different selective attention algorithm. Rather than relying on an assessment of the contribution of pixels to the
output of the last convolutional layer of a pretrained image
classification network, we propose performing sensitivity analyses at
all of the layers in the network and densely aggregating the resulting
information to guide selective attention. This approach is intended to
make use of both semantic information and spatial information by
considering channel-wise attention and spatial attention at every
convolutional layer in a pretrained classification CNN called the What Network. The channel-wise and
spatial sensitivity values for each layer are stacked to preserve
object location information implicit in the full network. The result
is used to guide selective attention, masking the input image before
it is presented to the Where Network for object detection. We also examined architectures for the Where Network appropriate for real time tasks. Finally, we compared this attention mechanism to human data.
\section{Algorithm}
\label{s:method}

\begin{figure*}[t]
  \centering
  \scalebox{0.95}{
  \begin{tabular}{c@{\hspace{0ex}}c@{\hspace{0ex}}c@{\hspace{0ex}}c@{\hspace{0ex}}c@{\hspace{0ex}}c@{\hspace{0ex}}c@{\hspace{0ex}}c@{\hspace{0ex}}c@{\hspace{0ex}}c@{\hspace{0ex}}}
  aeroplane&bike&bird&boat&bottle&bus&car&cat&chair&cow \\
    \includegraphics*[width=1.65cm,height=1.6cm]{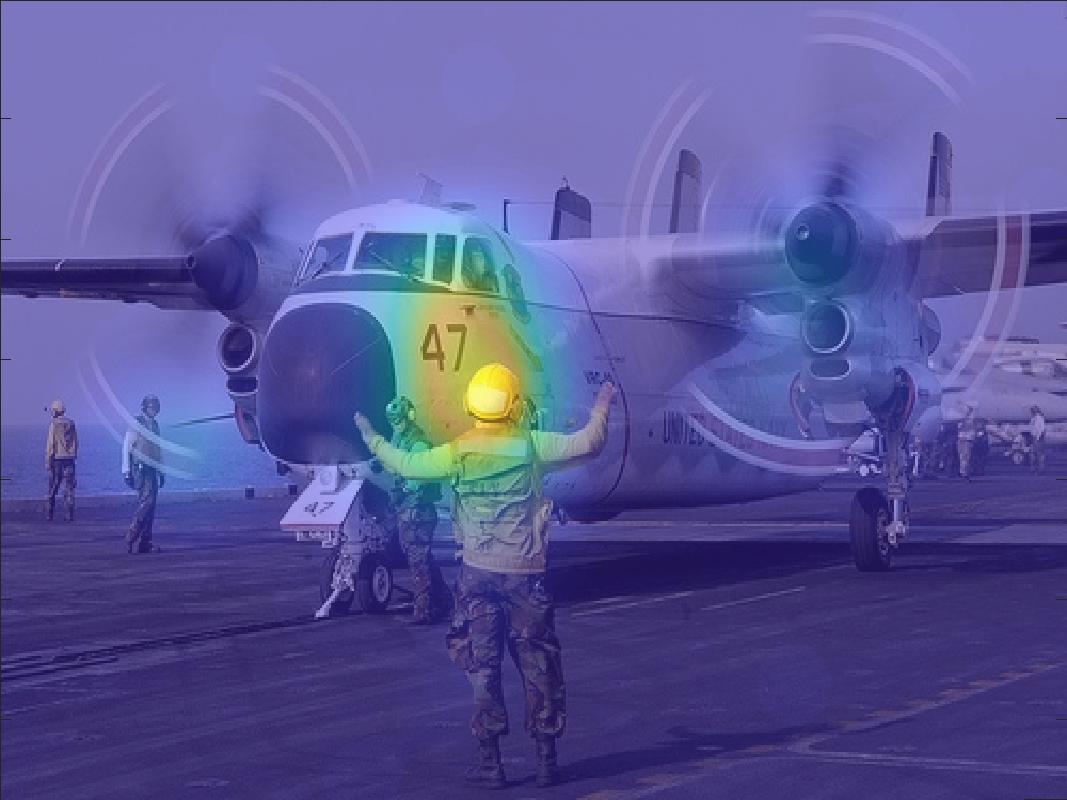}&
    \includegraphics*[width=1.65cm,height=1.6cm]{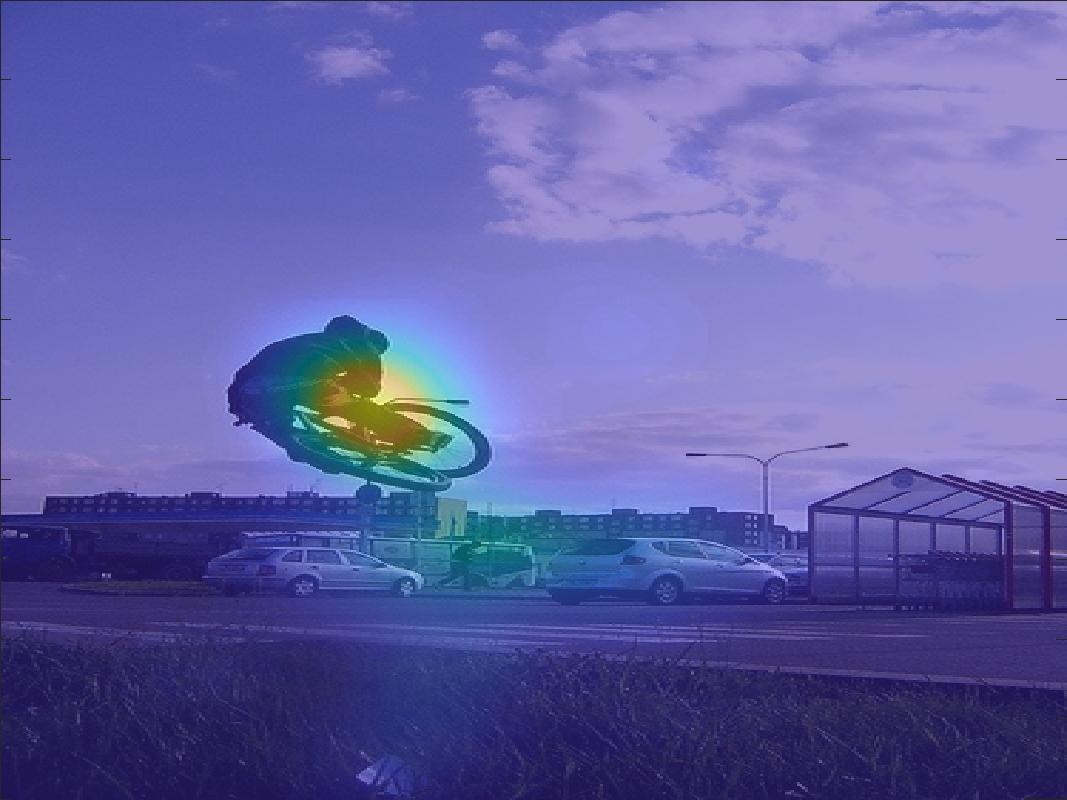}&
    \includegraphics*[width=1.65cm,height=1.6cm]{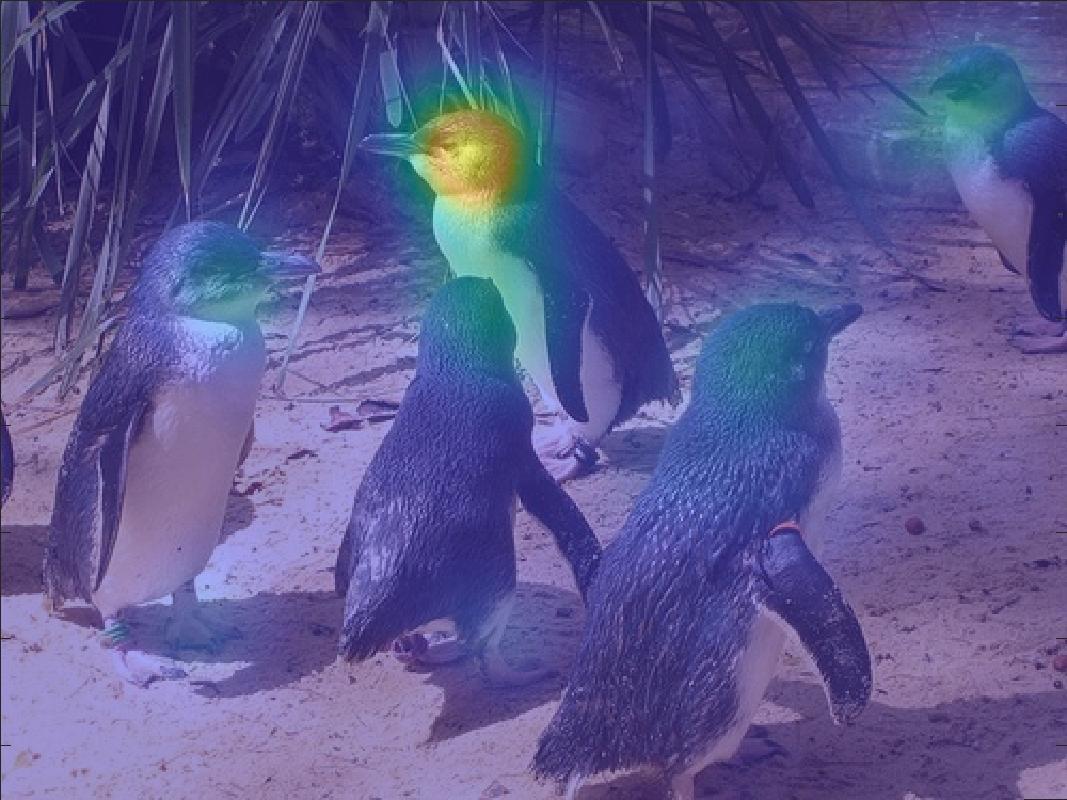}&
    \includegraphics*[width=1.65cm,height=1.6cm]{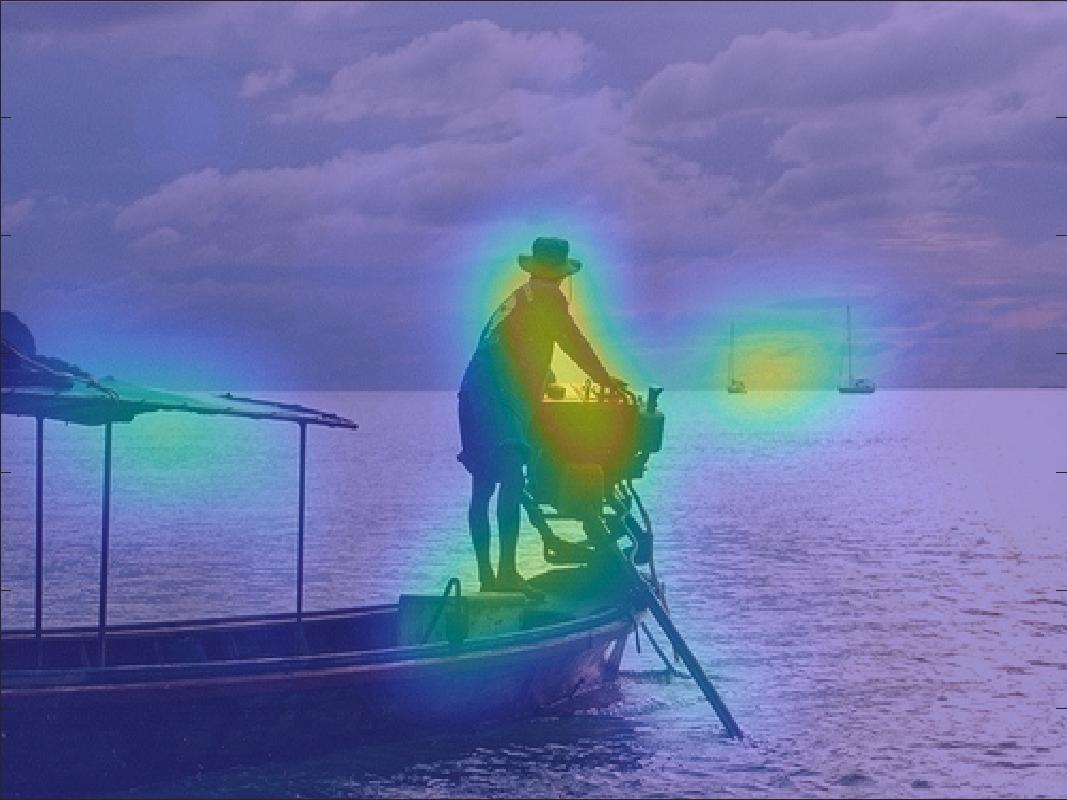}&
    \includegraphics*[width=1.65cm,height=1.6cm]{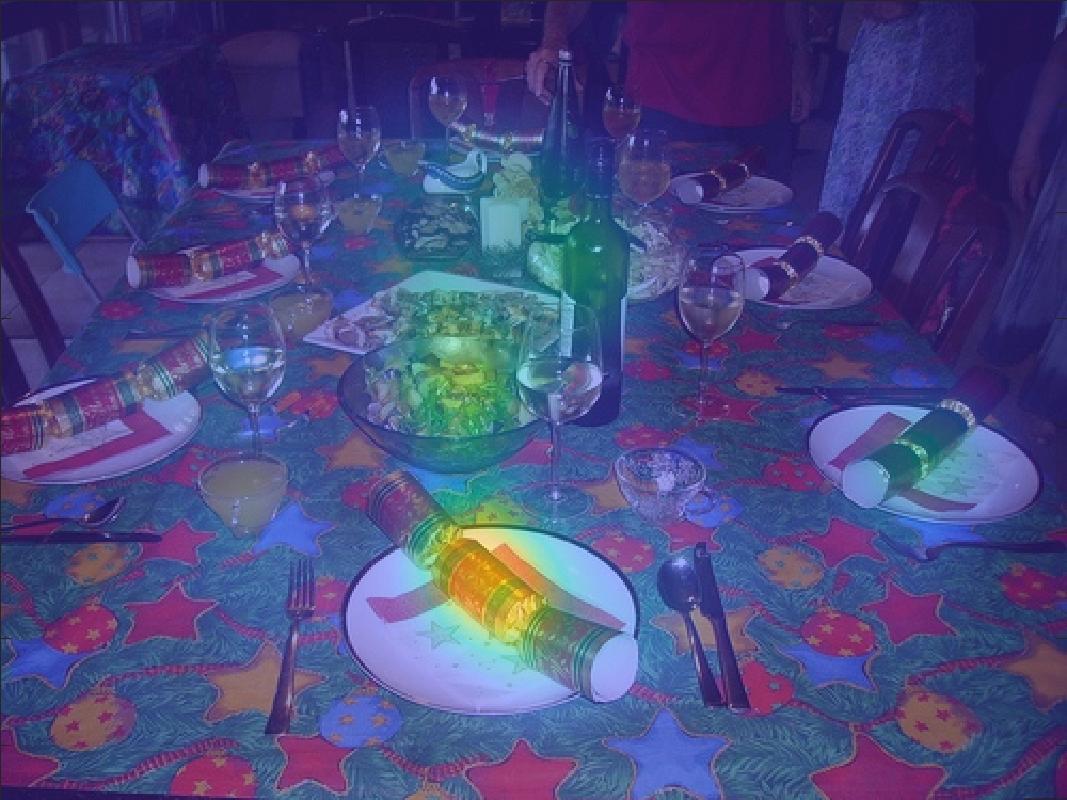}&
    \includegraphics*[width=1.65cm,height=1.6cm]{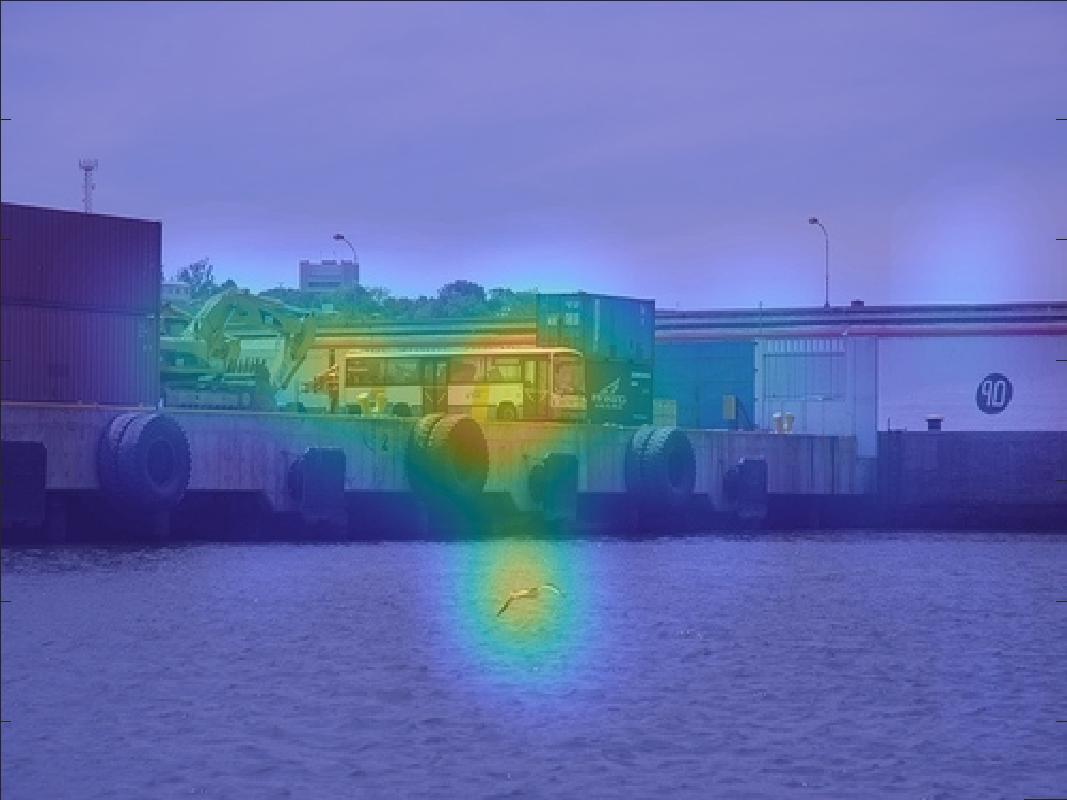}&
    \includegraphics*[width=1.65cm,height=1.6cm]{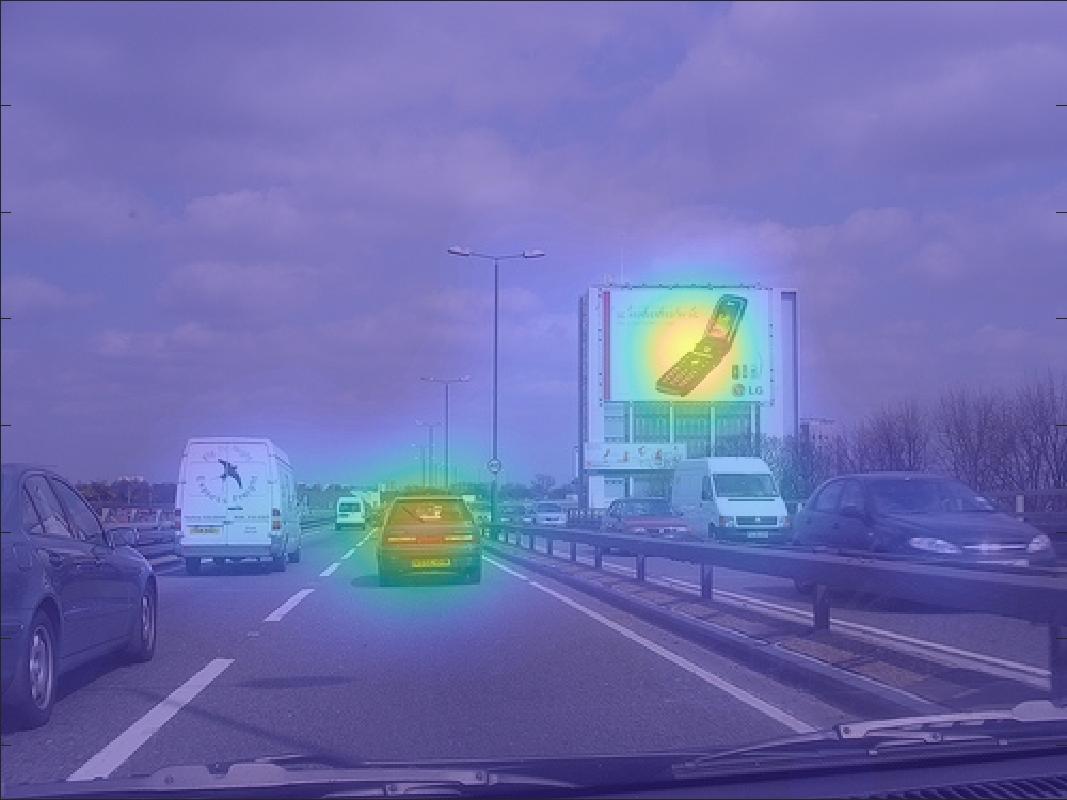}&
    \includegraphics*[width=1.65cm,height=1.6cm]{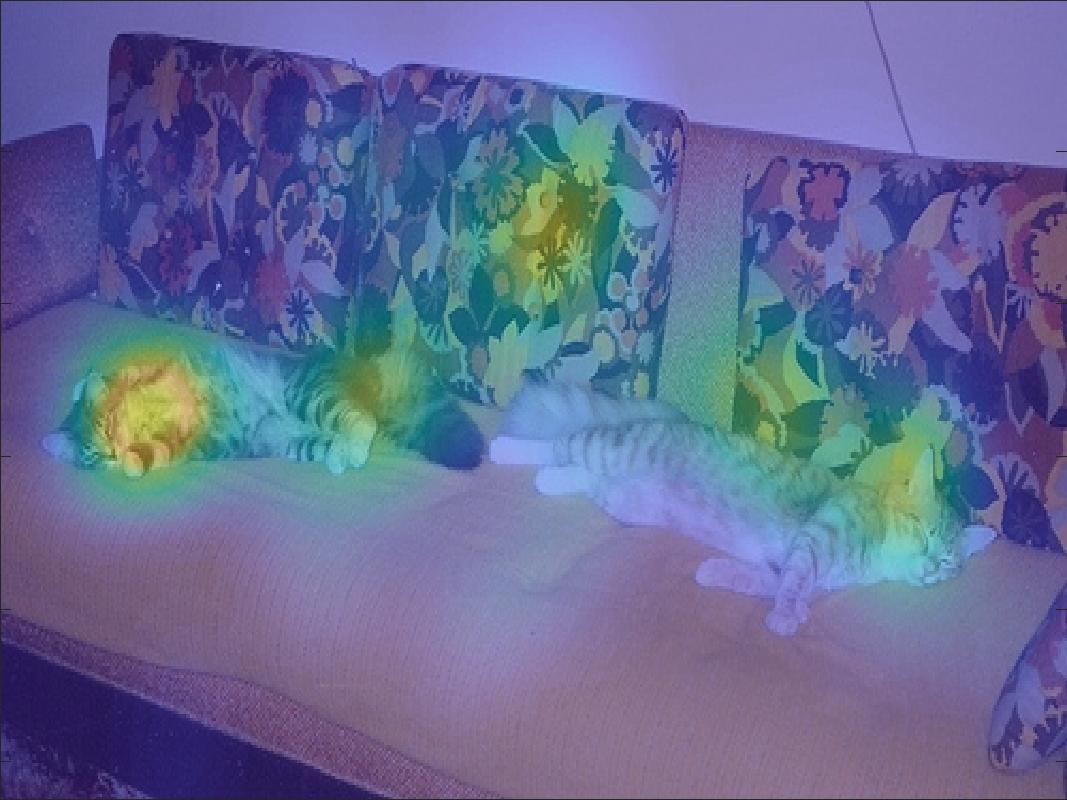}&
    \includegraphics*[width=1.65cm,height=1.6cm]{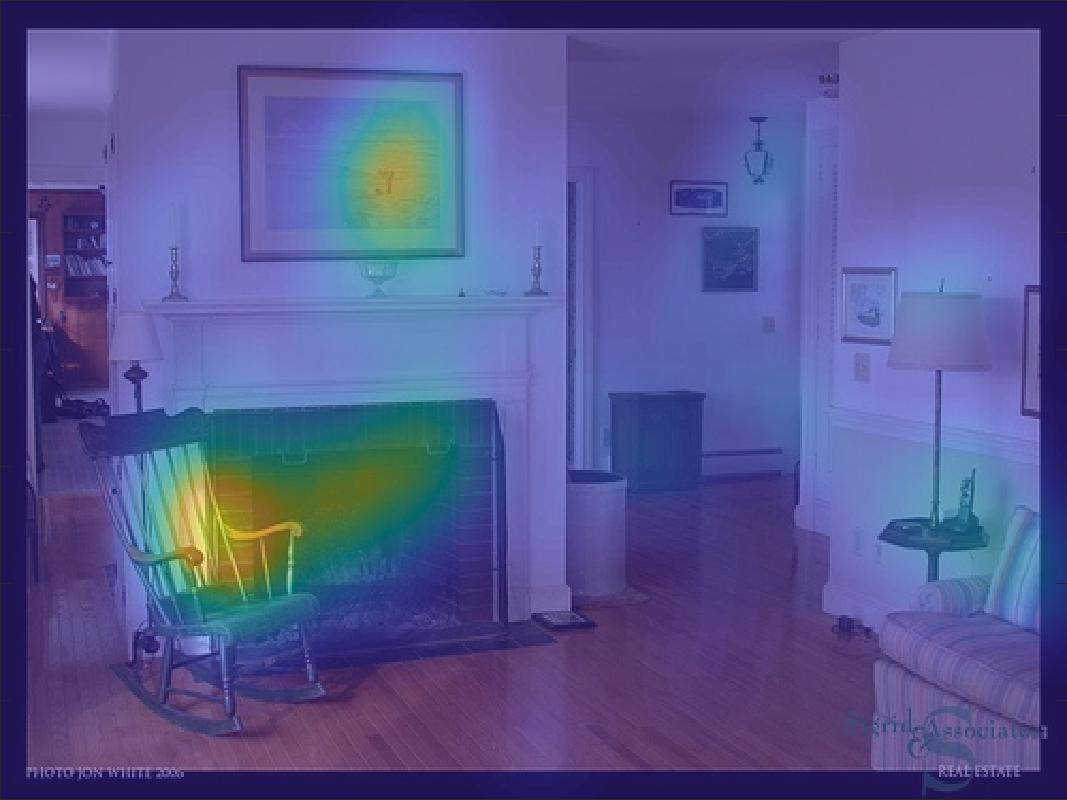}& 
   \includegraphics*[width=1.65cm,height=1.6cm]{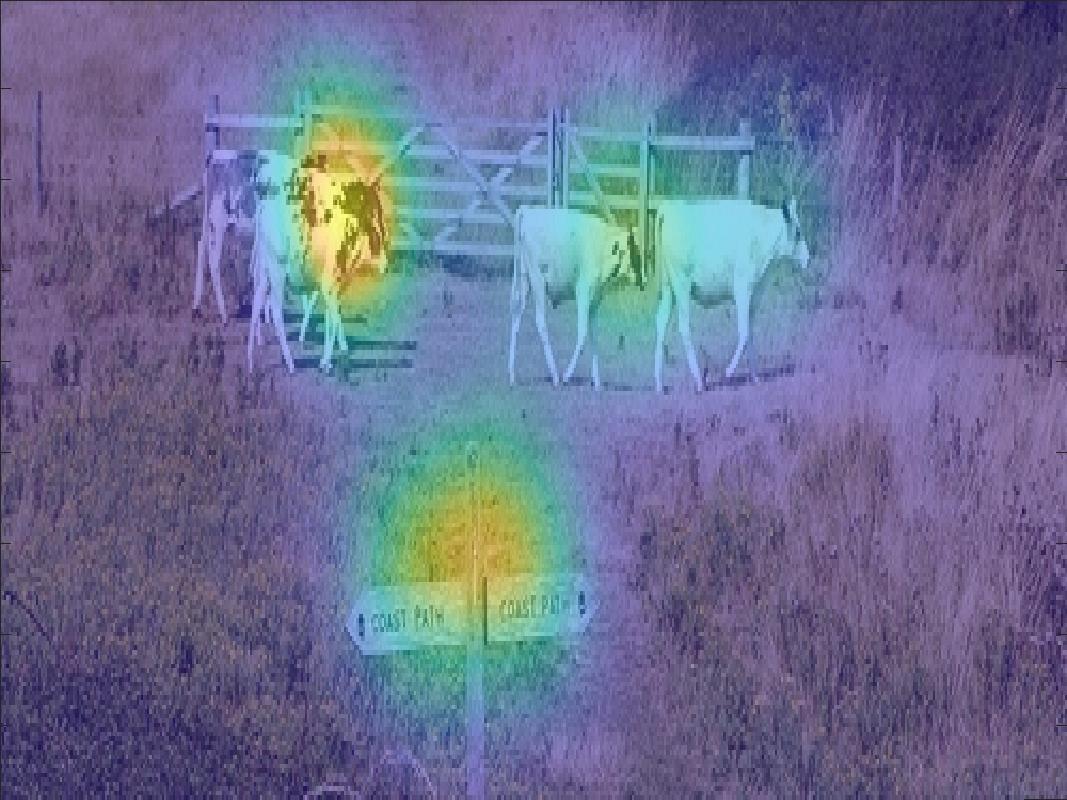}\\  
   \hline
   diningtable&dog&horse&motorbike&person&pottedplant&sheep&sofa&train&tv \\ 
    \includegraphics*[width=1.65cm,height=1.6cm]{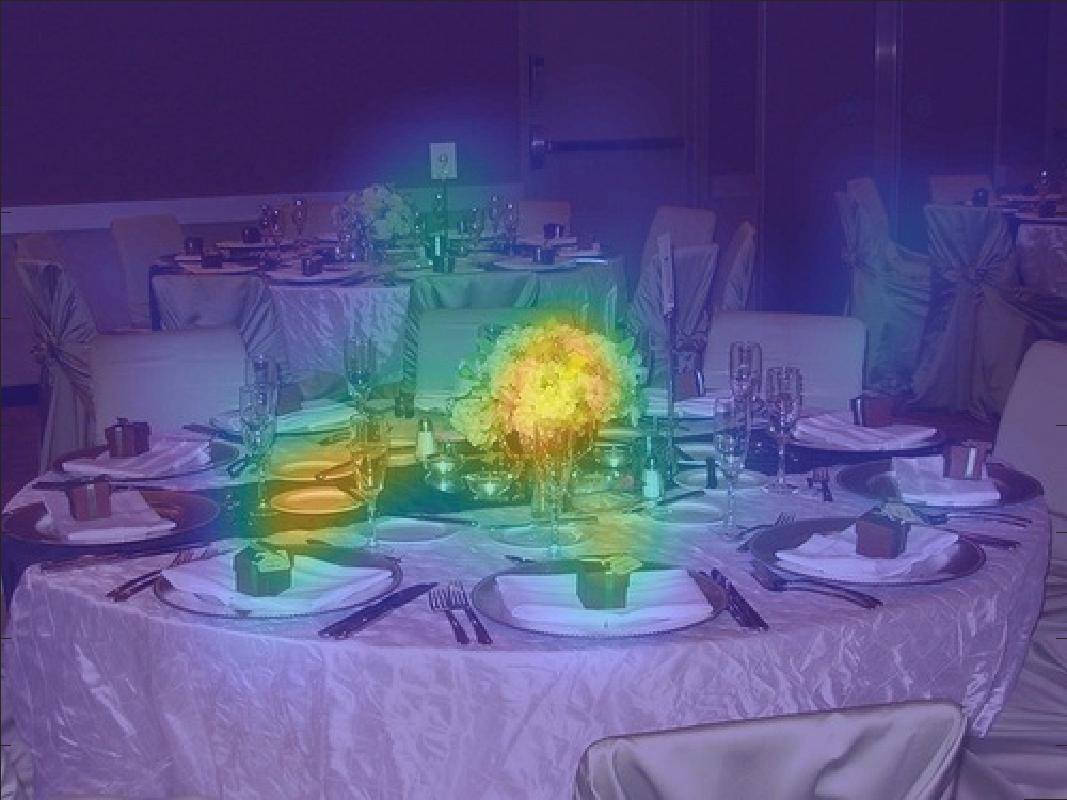}&
    \includegraphics*[width=1.65cm,height=1.6cm]{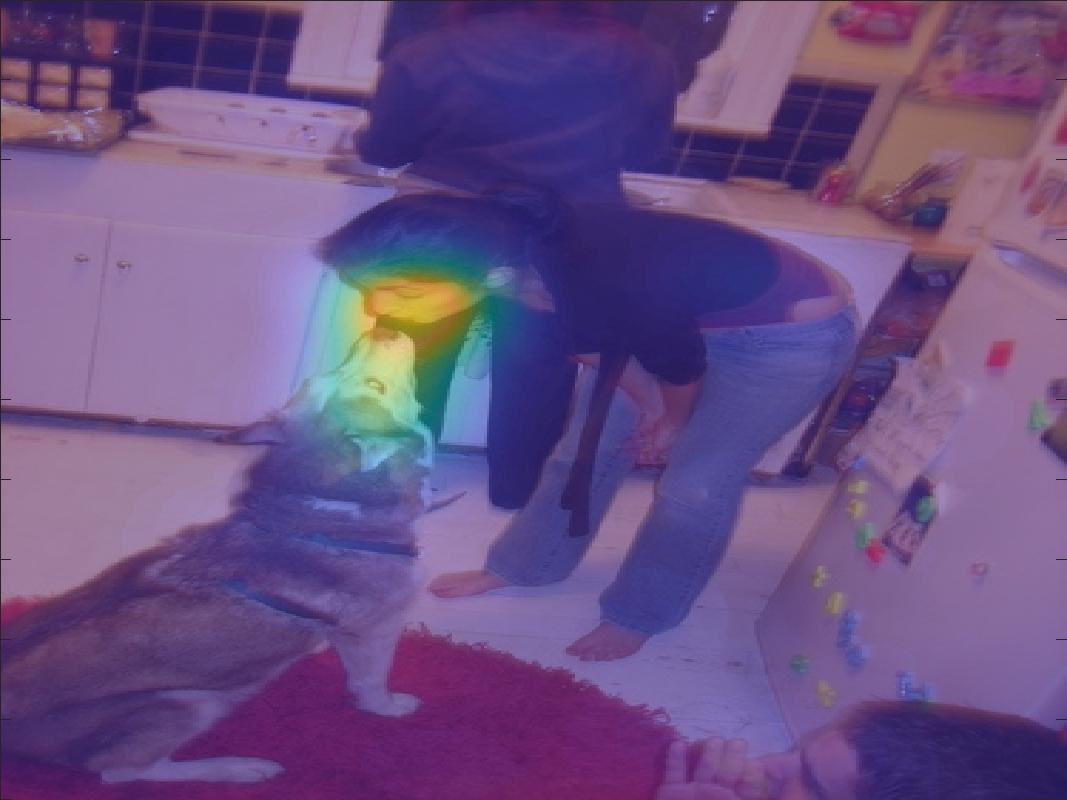}&
    \includegraphics*[width=1.65cm,height=1.6cm]{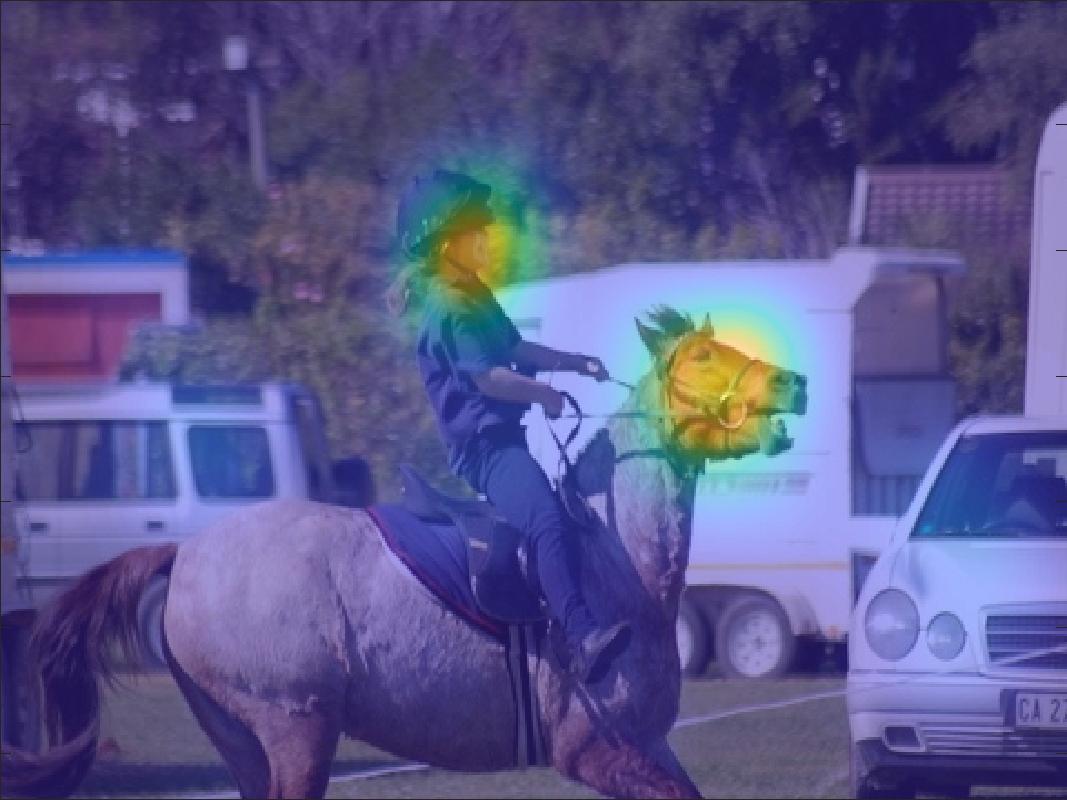}&
    \includegraphics*[width=1.65cm,height=1.6cm]{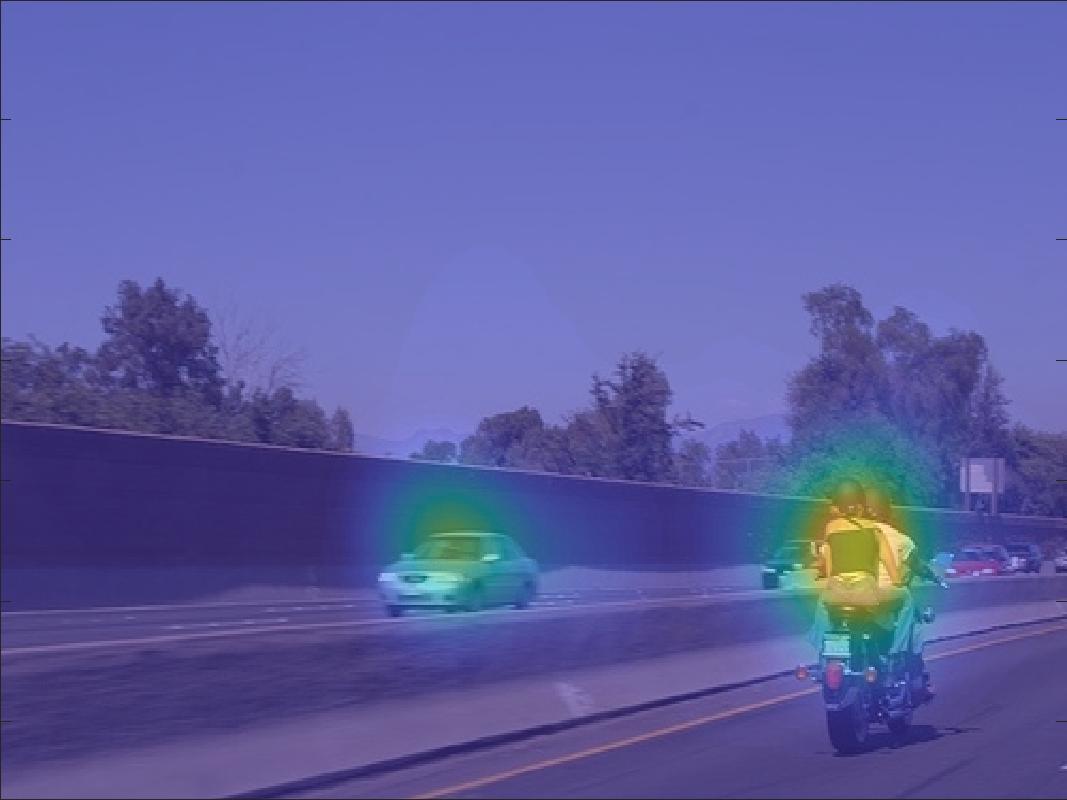}&
    \includegraphics*[width=1.65cm,height=1.6cm]{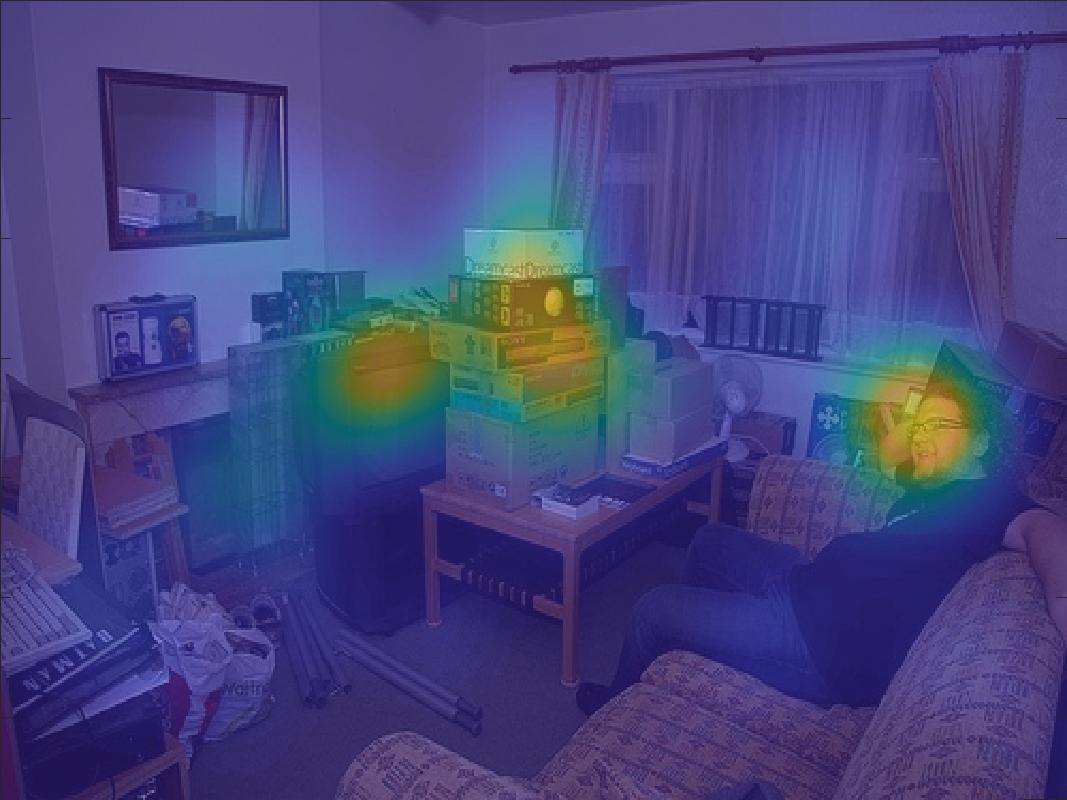}&
    \includegraphics*[width=1.65cm,height=1.6cm]{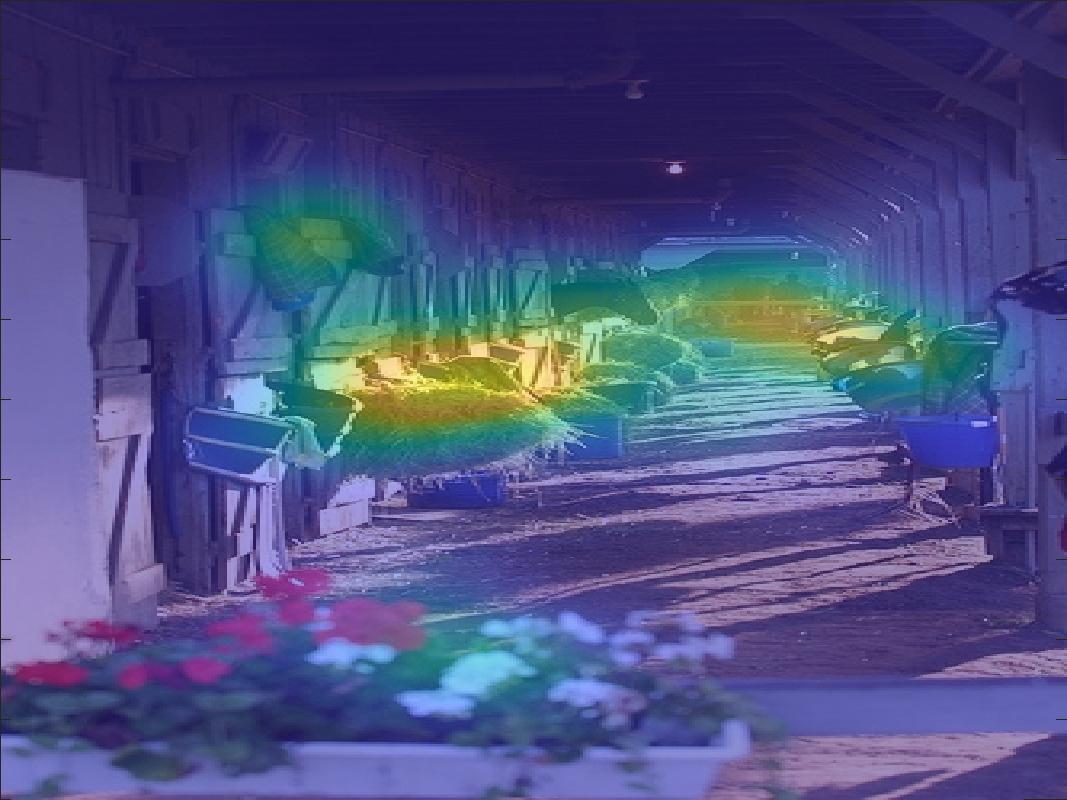}&
    \includegraphics*[width=1.65cm,height=1.6cm]{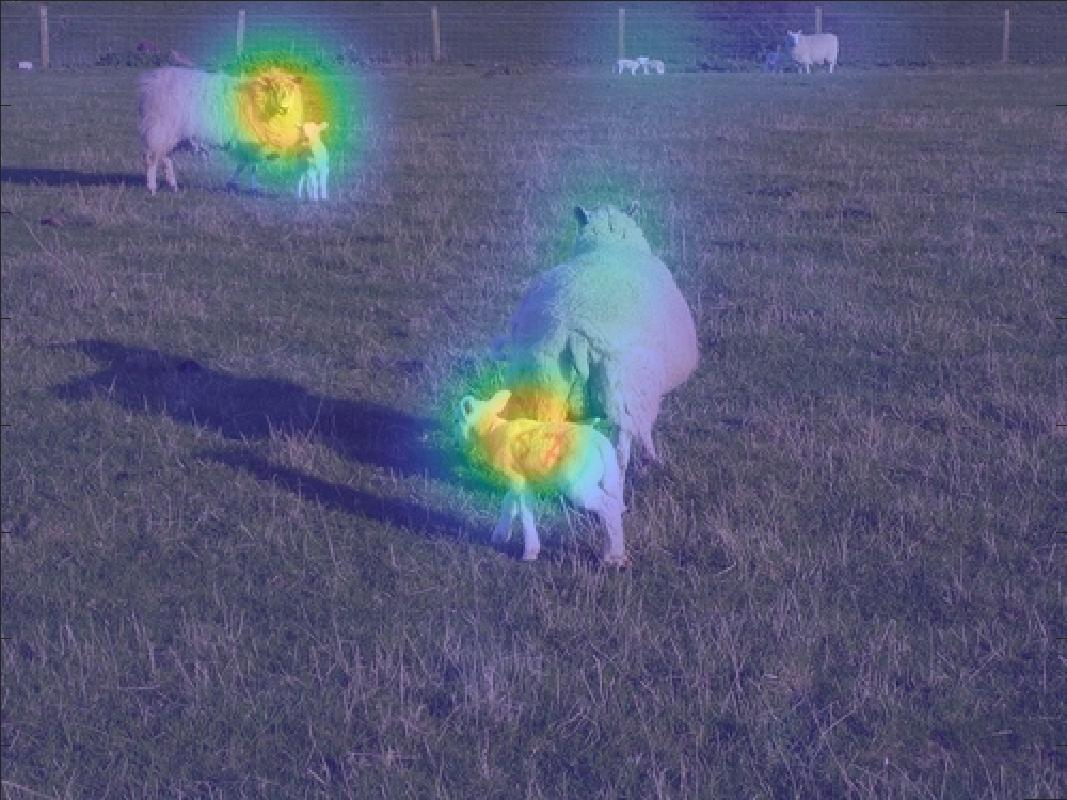}&
    \includegraphics*[width=1.65cm,height=1.6cm]{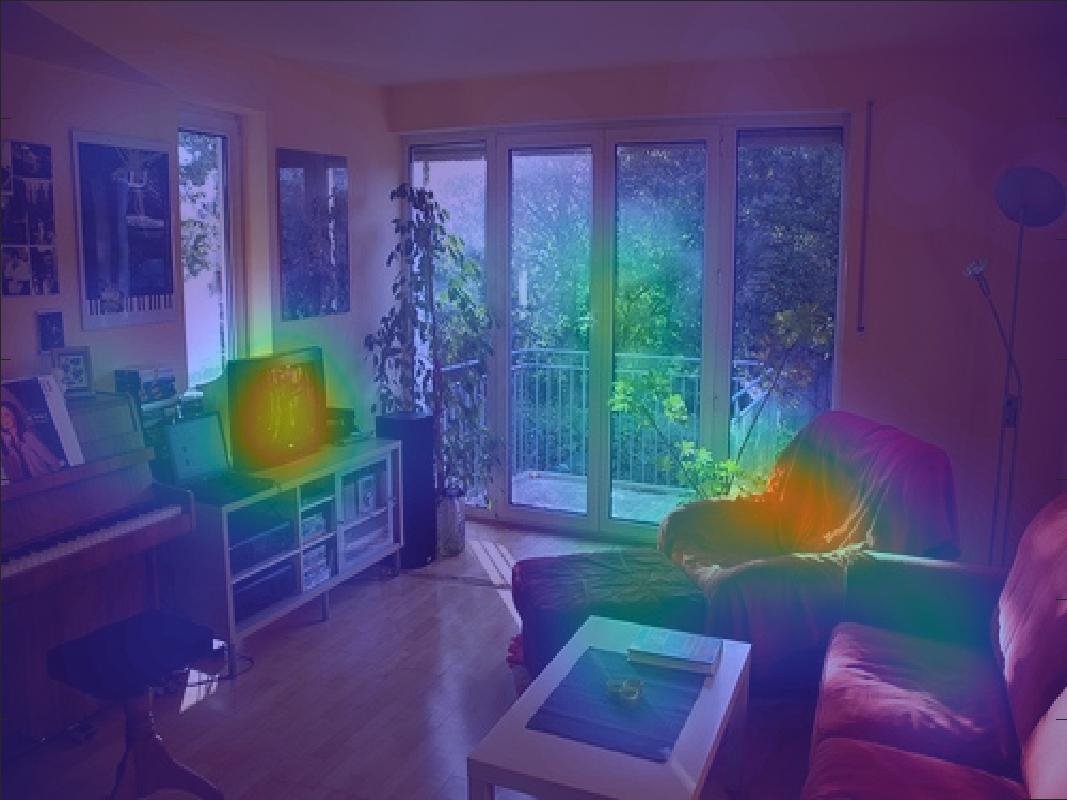}&
    \includegraphics*[width=1.65cm,height=1.6cm]{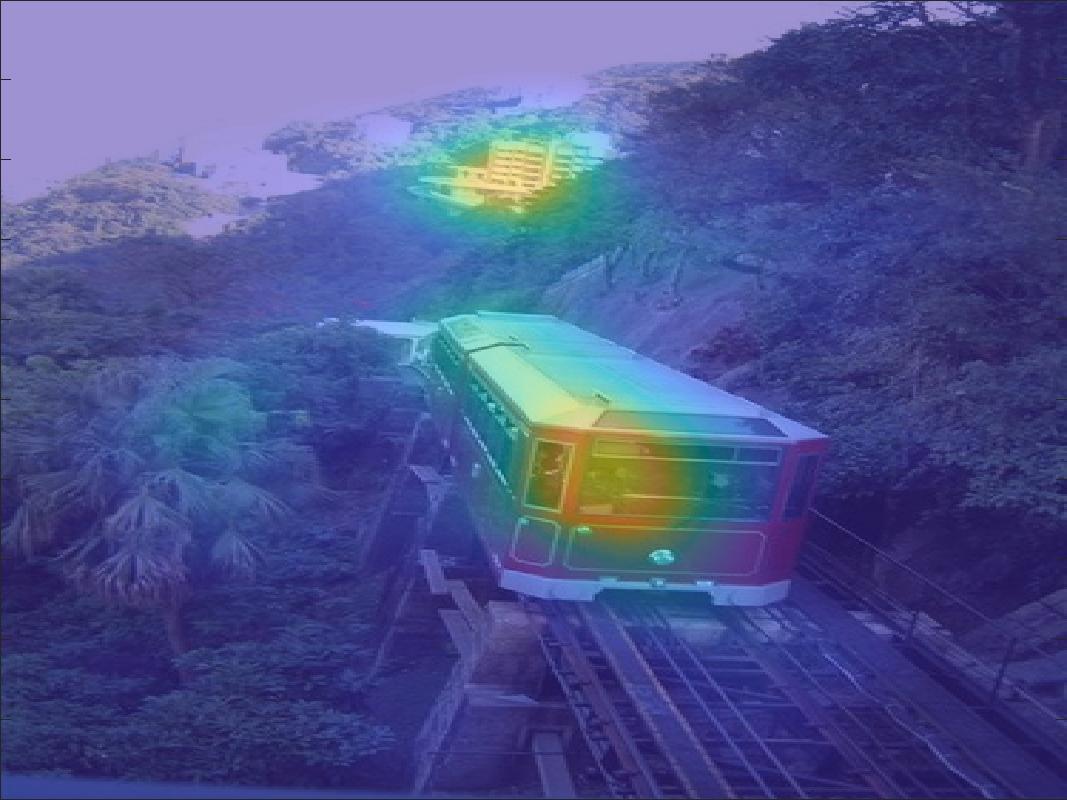}& 
   \includegraphics*[width=1.65cm,height=1.6cm]{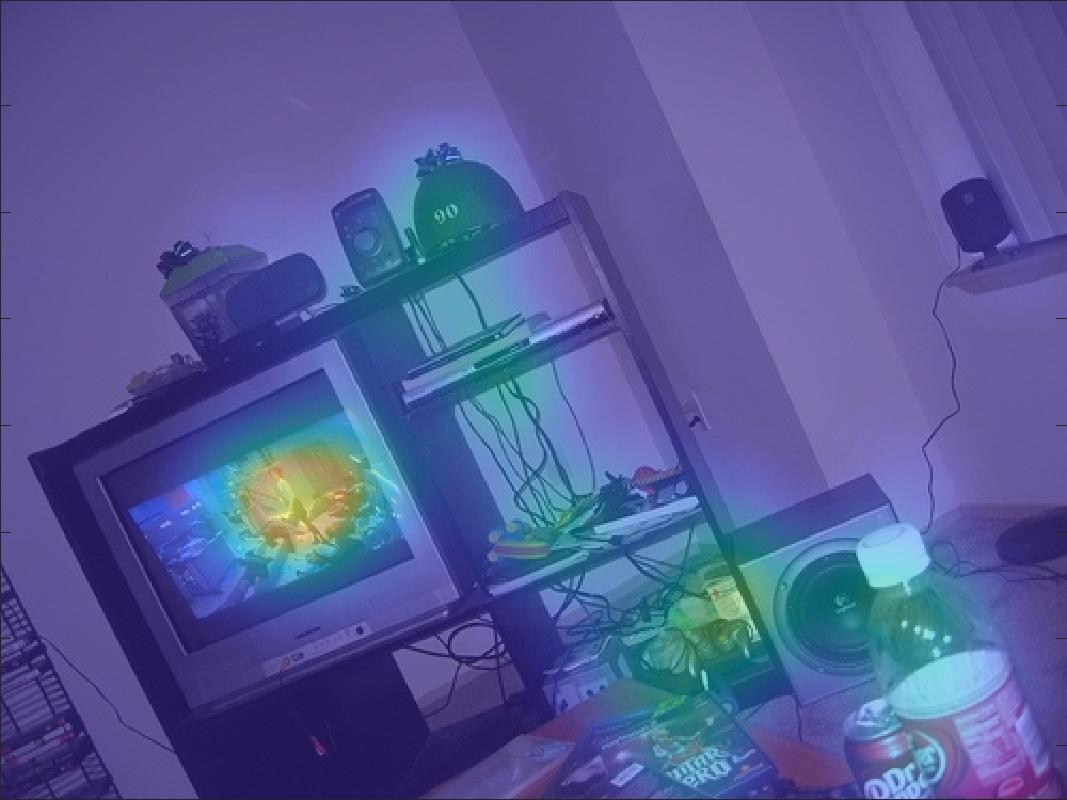}\\ 
  \end{tabular}}
  \caption{Example Human Eye Fixation Distributions on PASCAL VOC Images}
  \label{f:HA}
\end{figure*}

\textbf{Dense Attention of What Networks.}
Our investigations into WW-Nets used a pretrained image classification
network to guide selective attention. In our initial studies, for a
given input image, the sensitivity of activity late in the network to
each pixel was efficiently calculated, and regions containing low
sensitivity pixels were masked out. 
Our What Network has two major parts. First, we
calculated the saliency of image regions based on activity at layers
throughout the network, rather than only at the last convolutional
layer. We aggregated spatial and channel-wise sensitivity information
at each layer to produce layer-specific attention maps. Second, we
densely stacked the whole collection of attention maps, from all 
 layers to capture the most abstract features of the objects in the deep layers as well as obtaining the spatial locations of the objects in the shallow layers, in order to drive selective attention to image
regions. The architecture of WW-Nets is depicted in Figure~\ref{f:DCA}.\\
\subsection{Spatial \& Channel-wise Attention (SCA) Module}
\textbf{Spatial Attention.} Generally speaking, objects occupy only
parts of images, leaving background regions that can distract and
misinform object detection systems. Instead of considering all parts
of an image equally, spatial attention can focus processing on
foreground regions, supporting the extraction of features most
relevant for object class and object extent.

We represent a convolutional feature of layer $n$ by $f_n \in
\mathbb{R}^{W\times H \times C}$, where $W$ and $H$ are the spatial
dimensions of the rectangular layer and $C$ is the number of feature
channels in the layer. Spatial positions are specified by coordinate
pairs: $\mathbb{L}=\{(x,y)|x=1,2,\hdots,W; y=1,2,\hdots,H\}$.

For a layer in a pretrained image classification network, the
layer-specific spatial attention map is generated by the following
equation: 
\begin{equation}
A_n^s = W_n^s \odot f_n
\end{equation}
where $W_n$ are weights that indicate the importance of each spatial
location, across all of the convolutional filters, for the current image. We initially
calculate these weights based on the sensitivity of the Gestalt Total
(GT) activation of the network to the feature. The Gestalt Total is
calculated from the activation of the last convolutional layer,
$A_{\mbox{\it last}}$, as follows:
\begin{equation}
GT = \frac{1}{H \times W \times C}\sum_{i,j,k} A_{\mbox{\it last}}(i,j,k)
\end{equation}
GT is the activation over the most abstract features in the last convolutional layer.
The sensitivity of GT to a feature at layer $n$ is the following:
\begin{equation}
\begin{split}
G_n = \frac{\partial GT}{\partial f_n};\   
W_n^u(x,y) = \sum_c^C G_n (x,y,c)
\end{split}
\end{equation}
where $W_n^u$ is not normalized (i.e., the weights are not in the
$[0,1]$ range). To normalize the weights for each location,
$\ell$, we apply a softmax operation to the weights spatially:
\begin{equation}
W_n^s( l) = \frac{\exp (W_n^u(l))}{\sum_{ i \in \mathbb{L}} \exp(W_n^u(i))}
\end{equation}
%

\noindent where $W_n^s(l)$ denotes the weight for location $l$ in layer $n$.\\[-2mm]

\textbf{Channel-Wise Attention.} The spatial attention calculation
assigns weights to spatial locations, which addresses the problem of
distractions from background regions. There is another way in which
distractions can arise, however. Specific channels at a given layer
can be distracting. When dealing with convolutional features, most
existing methods treat all channels without distinction. However,
different channels are often different in their relevance for objects
of specific classes. Here, we introduce a channel-wise attention
mechanism that assigns larger weights to channels to which the $GT$ is
sensitive, given the currently presented image. Incorporating these
channel-wise attentional weights are intended to reduce this kind of
distracting interference.

For channel-wise attention, we unfold $f_n$ as $f =[f_n^1,f_n^2,
\hdots,f_n^C]$, where $f_n^i \in \mathbb{R}^{W \times H}$ is the
i\textsuperscript{th} slice of $f_n$, and $C$ is the total number of
channels. The goal is to calculate a weight, $W_n^{c}$, to scale the
convolutional features according to a channel-specific assessment of
relevance:
\begin{equation}
A_n^c = W_n^c \cdot f_n
\end{equation}
Computing $W_n^c$ is facilitated by the fact that we already have the
sensitivities, $G_n$. Thus, an initial value for the weights can be
had by setting $\hat{W}_n^{c}(c) = \sum_{x \in W, y \in H}
G_n(x,y,c)$. These weights can be normalized to the $[0,1]$ range
using the softmax function:
\begin{equation}
W_n^c( c ) = \frac{\exp (\hat{W}_n^{c}(c))}{\sum_{ i \in C } \exp(\hat{W}_n^{u}(i))}
\end{equation}

\noindent These are the final channel-wise weights for layer $n$.\\[-2mm]

\textbf{Dense Attention Maps.}
Given the spatial attention weights and the channel-wise attention
weights, an attention weighted feature for layer $n$ is calculated as:
\begin{equation}
f_n^{SCA} = A_n^c \cdot f_n + A_n^s \odot f_n
\end{equation}    
where $f_n^{SCA} \in \mathbb{R}^{W \times H \times C}$. These weighted
features are concatenated across layers, incrementally from the last
layer to the first, but this is done after up-scaling lower spatial
resolution layers. For the last layer, $m$, the map is simply the
weighted features: $A_{m} = f_{m}^{SCA}$. For earlier layers:
\begin{equation}
\begin{split}
A_{i} = [UP(f_{m}^{SCA}),UP(f_{m-1}^{SCA}),\hdots,f_i^{SCA}] \\
i=\{1,\hdots,m-1\}
\end{split}
\end{equation}
The result is dense combination maps that are intended to
incorporate both semantic information from the late layers and spatial
information from the early layers. These attention maps are then
combined to make a single map for the whole image. Since each layer
can have a different number of channels, we simplify this aggregation
by averaging each layer's attention map across channels, tranforming
each $A_{i}$ into a $W \times H$ matrix. The aggregated attention maps
at each layer are computed as:
\begin{equation}
Att_{i} = A_{m}^{SCA} \oplus A_{m-1}^{SCA} \oplus \hdots \oplus A_i^{SCA}
\end{equation} 
The process of spatial \& channel wise attention, along with dense connections among them, is depicted in Figure~\ref{f:DCA}.

The final aggregated attention map, at the first layer, is
$Att_{1}$. We smooth this $W \times H$ map by convolving it with a
Gaussian filter, and then we threshold the result. This produces a
binary mask specifying regions of relevance. The original image is
multiplied by this mask to produce the input to the Where Network.

It is important to note that the attention weights ($W_n^s$ and
$W_n^c$) are not learned as part of a training process. We begin with
a pretained image classification network for the Ventral Network, and
the attention weights are efficiently calculated for each layer when a
given image is presented to that network.\\[-2mm]

\textbf{Where Network.}
Blanking irrelevant regions in the image substantially reduces the
space of candidate regions to consider during object detection. In the
Where Network, the masked image is provided as input to a deep CNN
trained to propose regions of interest with anchor boxes, process the
contents of those regions, and output both class labels and bounding
box coordinates. This is similar to the approach used by
Faster-RCNN~\cite{fasterrcnn}. The Where Net is trained using a
dataset of images that are annotated with both ground-truth class
labels and ground-truth bounding boxes. Network parameters are
selected so as to minimize a combination of the classification loss
and the regression loss arising from the output of bounding box
coordinates:
\begin{equation}
\begin{split}
L(p_i,t_i) = \frac{1}{N_{cls}}\sum_i L_{cls}(p_i,p_i^{\ast}) \\
+\lambda \frac{1}{N_{reg}} \sum_i p_i^{\ast} L_{reg}(t_i,t_i^{\ast})
\end{split}
\label{eq:objFunction}
\end{equation}
where $i$ is the index of an anchor box appearing in the current
training mini-batch and $p_i$ is the predicted probability of anchor
$i$ containing an object of interest. The ground-truth label
$p_i^{\ast}$ is $1$ if anchor $i$ is positive for object presence and
it is $0$ otherwise. The predicted bounding box is captured by the 4
element vector $t_i$, and $t_i^{\ast}$ contains the coordinates of the
ground-truth bounding box associated with a positive anchor. The two
components of the loss function are normalized by $N_{cls}$ and
$N_{reg}$, and they are weighted by a balancing parameter,
$\lambda$. In our current implementation, the classification loss term
is normalized by the mini-batch size (i.e., $N_{cls} = 32$) and the
bounding box regression loss term is normalized by the number of
anchor locations (i.e., $N_{reg} \approx 2,400$). We set $\lambda =
10$, making the two loss terms roughly equally weighted (due to
differences in scale).

It is worth noting that our general approach could easily incorporate
other object detection algorithms for the Where Net. The only
requirement is that the object detection algorithm must be able to
accept masked input images. For the main results presented in this paper,
we have used a region proposal based approach, due to the high
accuracy values reported for these methods in the literature. Having
the What Net reduce the number of proposed regions was expected
to speed the object detection process and also potentially improve
accuracy by removing from consideration irrelevant portions of the
image. As discussed in the next section, we also trained YOLO as an alternative Where Net for achieving real time performance.
\section{Experimental Results}
\label{S:experiments}
\textbf{Experiment Design and Implementation}.
We evaluated the WW-Nets object detection model on PASCAL VOC 2007~\cite{voc2007}, PASCAL VOC 2012~\cite{voc2012}, and COCO datasets~\cite{coco}.
We also
compared the attention model to human performance. Source code and human selective attention heatmaps will be made publicly available.

The PASCAL VOC 2007 dataset has 20 classes and $9,963$ images which
have been equally split into a training/validation set and a test set.
The PASCAL VOC 2012 dataset contains $54,900$ images from 20 different
categories, and it has been split approximately equally into a
training/validation set and a test set.  For PASCAL VOC 2007, we
conducted training on the union of the VOC 2007 trainval set and the
VOC 2012 trainval set, and we evaluated the results using the VOC 2007
test set. (This regimen is standard practice for these datasets.) For
PASCAL VOC 2012, we performed training on its trainval set, and we
evaluated the result on its test set. To evaluate performance, we used
the standard mean average precision (mAP) measure. We report mAP
scores using IoU thresholds at $0.5$ for PASCAL datasets and we used mAP with both IOU thresholds at $0.5$ and $0.75$ for the COCO dataset.

For networks with $224 \times 224$ image inputs, using PASCAL VOC, we
trained the model with a mini-batch size of 1 due to GPU memory
constraints. We started the learning rate at $3 \times 10^{-4}$ for
the first $900,000$ iterations. We then decreased it to $3 \times 10^{-5}$
until iteration $1,200,000$. Then, we decreased it to $3 \times 10^{-6}$
until iteration $2,000,000$. In all cases, we used a momentum optimizer
value of $0.9$. 
\begin{table*}[t]
\begin{center}
\caption{PASCAL VOC 2007 Test Detection Results. Note that the minimum
         dimension of the input image for Faster and R-FCN is 600, and
         the speed is less than 10 frames per second. SSD300 indicates
         the input image dimension of SSD is $300 \times 300$. Large
         input sizes can lead to better results, but this increases
         running times. All models on the union of the
         trainval set from VOC 2007 and VOC 2012 and tested on the VOC
         2007 test set, except for the model labeled $\text{
         WW-Nets}^{\ast}$  which has been trained on the trainval set
         from VOC 2007 and tested on 2007 test set. Also, $\text{
         WW-Nets}^{+}$ is trained on the union of trainval set from VOC 2007 and VOC 2012 and ImageNet and tested on the VOC
         2007 test set.
          \label{t:voc2007} } 
\scalebox{0.7}{
 \begin{tabular}{c|c|c|cccccccccccccccccccc} 
 \hline
 Method & Network & mAP & areo& bike&bird&boat&bottle&bus&car&cat&chair&cow&table&dog&horse&mbike&person&plant&sheep&sofa&train&tv \\  
 \hline\hline
Faster \cite{fasterrcnn}& VGG& 73.2& 76.5& 79& 70.9& 65.5& 52.1& 83.1& 84.7& 86.4& 52& 81.9& 65.7& 84.8& 84.6& 77.5& 76.7& 38.8& 73.6& 73.9 &83& 72.6 \\ 
ION \cite{bell} & VGG& 75.6& 79.2& 83.1& 77.6& 65.6& 54.9& 85.4& 85.1& 87& 54.4& 80.6& 73.8& 85.3& 82.2& 82.2& 74.4& 47.1& 75.8& 72.7& 84.2& 80.4\\
Faster \cite{resNet}& Residual-101& 76.4& 79.8& 80.7& 76.2& 68.3& 55.9& 85.1& 85.3& 89.8& 56.7& 87.8& 69.4& 88.3& 88.9& 80.9& 78.4& 41.7& 78.6& 79.8& 85.3& 72\\
MR-CNN \cite{multi-region}& VGG& 78.2& 80.3& 84.1& 78.5& 70.8& 68.5& 88& 85.9& 87.8& 60.3& 85.2& 73.7& 87.2& 86.5& 85& 76.4& 48.5& 76.3& 75.5& 85& 81\\
R-FCN \cite{region-based}& Residual-101& 80.5& 79.9& 87.2& 81.5& 72& 69.8& 86.8& 88.5& 89.8& 67& 88.1& 74.5& 89.8& 90.6& 79.9& 81.2& 53.7& 81.8& 81.5& 85.9& 79.9\\
\hline
SSD300 \cite{ssd} &VGG& 77.5& 79.5& 83.9& 76& 69.6& 50.5& 87& 85.7& 88.1& 60.3& 81.5& 77& 86.1& 87.5& 83.9& 79.4& 52.3& 77.9& 79.5& 87.6& 76.8 \\
SSD512 \cite{ssd}& VGG& 79.5& 84.8& 85.1& 81.5& 73& 57.8& 87.8& 88.3& 87.4& 63.5& 85.4& 73.2& 86.2& 86.7& 83.9& 82.5& 55.6& 81.7& 79& 86.6& 80\\
\hline
DSSD321 \cite{dssd}& Residual-101& 78.6& 81.9& 84.9& 80.5& 68.4& 53.9& 85.6& 86.2& 88.9& 61.1& 83.5& 78.7& 86.7& 88.7& 86.7& 79.7& 51.7& 78& 80.9& 87.2& 79.4\\
DSSD513 \cite{dssd}& Residual-101& 81.5& 86.6& 86.2& 82.6& 74.9& 62.5& 89& 88.7& 88.8& 65.2& 87& 78.7& 88.2& 89& 87.5& 83.7& 51.1& 86.3& 81.6& 85.7& 83.7\\
\hline
STDN300 \cite{scale-transfer} &DenseNet-169&78.1& 81.1& 86.9& 76.4& 69.2& 52.4& 87.7& 84.2& 88.3& 60.2& 81.3& 77.6& 86.6& 88.9& 87.8& 76.8& 51.8& 78.4& 81.3& 87.5& 77.8\\
STDN321 \cite{scale-transfer}& DenseNet-169& 79.3& 81.2& 88.3& 78.1& 72.2& 54.3& 87.6& 86.5& 88.8& 63.5& 83.2& 79.4& 86.1& 89.3& 88.0& 77.3& 52.5& 80.3& 80.8& 86.3& 82.1\\
STDN513 \cite{scale-transfer}& DenseNet-169& 80.9& 86.1& 89.3& 79.5& 74.3& 61.9& 88.5& 88.3& 89.4& 67.4& 86.5& 79.5& 86.4& 89.2& 88.5& 79.3& 53.0& 77.9& 81.4& 86.6& 85.5\\
\hline
VDNet \cite{VDNet}
 & Resnet-101&86.2& 95.8& 98.1& 98.4& 65.1& 94.6& 90.1& 96.2& 71.7& 72.3& 54.6& 97.9& 95.6& 89.2& 90.1& 93.2& 69.1& 89.2& 82.1& 93.4.6& 74.0\\
\hline
$\text{WW-Nets}^{\ast}$& YOLONet&61.4& 66.5& 69.5& 61.9& 39.2& 42.8& 69.2& 65.3& 79.3& 44.2& 57.3& 54.8& 72.9& 77.1& 69.1& 72.4& 34.2& 49.5& 63.1& 76.1& 65.2\\
  WW-Nets& YOLONet&63.7& 80.1& 73.2& 54.1& 47.2& 43.1& 74.5& 73.2& 78.6& 42.1& 62.8& 57.3& 74.9& 77.7& 73.6& 73.2& 30.2& 53.1& 64.1& 75.9& 65.9\\
   WW-Nets & Resnet-101&\textbf{86.7}& 95.8& 98.4& 98.2& 66.4& 94.6& 90.2& 96.1& 71.2& 72.8& 54.9& 97.9& 95.6& 89.6& 90.3& 93.2& 69.6& 89.2& 82.1& 93.2& 74.1\\
$\text{WW-Nets}^{+}$ & Resnet-101&\textbf{88.1}& 94.1& 97.7& 98.9& 67.7& 94.5& 92.1& 95.9& 72.1& 73.2& 52.2& 98.0& 96.3& 87.9& 91.4& 91.8& 70.6& 91.1& 81.9& 95.1& 72.6\\
\end{tabular}}
\end{center}
\end{table*}
\begin{table*}[bt]
\caption{PASCAL VOC 2012 Test Detection Results.$\text{
         WW-Nets}^{+}$ is trained on the union of trainval VOC 2012 and ImageNet.Note that the
         performance of WW-Nets is about $6\%$ better than baseline Faster-RCNN. \label{t:voc2012}} 
\begin{center}
\scalebox{0.75}{
 \begin{tabular}{c|c|cccccccccccccccccccc} 
 \hline
Method & mAP& aero& bike& bird& boat& bottle& bus& car& cat& chair& cow& table& dog& horse& mbike& person& plant& sheep& sofa& train& tv \\
 \hline\hline
HyperNet-VGG \cite{hyperNet}& 71.4& 84.2& 78.5& 73.6& 55.6& 53.7& 78.7& 79.8& 87.7& 49.6& 74.97 &52.1& 86.0& 81.7& 83.3& 81.8& 48.6& 73.5& 59.4& 79.9& 65.7\\
HyperNet-SP \cite{hyperNet}& 71.3& 84.1& 78.3& 73.3& 55.5& 53.6& 78.6& 79.6& 87.5& 49.5& 74.9& 52.1& 85.6& 81.6& 83.2& 81.6& 48.4& 73.2& 59.3& 79.7& 65.6\\
Fast R-CNN $+$ YOLO \cite{yolo}& 70.7& 83.4& 78.5& 73.5& 55.8& 43.4& 79.1& 73.1& 89.4& 49.4& 75.5& 57.0& 87.5& 80.9& 81.0& 74.7& 41.8& 71.5& 68.5& 82.1& 67.2 \\
MR-CNN-S-CNN \cite{MR-CNN-S-CNN}& 70.7& 85.0& 79.6& 71.5& 55.3& 57.7& 76.0& 73.9& 84.6& 50.5& 74.3& 61.7& 85.5& 79.9& 81.7& 76.4& 41.0& 69.0& 61.2& 77.7& 72.1 \\
Faster R-CNN \cite{fasterrcnn}& 70.4& 84.9& 79.8& 74.3& 53.9& 49.8& 77.5& 75.9& 88.5& 45.6& 77.1& 55.3& 86.9& 81.7& 80.9& 79.6& 40.1& 72.6& 60.9& 81.2& 61.5 \\
NoC \cite{NoC}& 68.8& 82.8& 79.0& 71.6& 52.3& 53.7& 74.1& 69.0& 84.9& 46.9& 74.3& 53.1& 85.0& 81.3& 79.5& 72.2& 38.9& 72.4& 59.5& 76.7& 68.1\\
 VDNets  \cite{VDNet} 
 & 73.2& 85.1& 82.4& 73.6& 57.7& 61.2& 79.2& 77.1& 85.5& 54.9& 79.8& 61.4& 87.1& 83.6& 81.7& 77.9& 45.6& 74.1& 64.9& 80.3& 73.1\\
\hline
WW-Nets & \textbf{74.1}& 85.7& 82.2& 74.1& 55.9& 60.9& 79.8& 76.4& 83.5& 56.9& 78.4& 63.4& 88.4& 83.9& 81.3& 77.1& 46.5& 74.6& 65.3& 80.1& 72.9\\
$\text{WW-Nets}^{+}$ & \textbf{76.8}& 87.2& 81.9& 75.7& 56.3& 61.9& 76.2& 76.1& 85.1& 55.7& 79.8& 62.5& 87.2& 82.8& 82.4& 79.5& 43.2& 73.1& 64.1& 81.7& 73.2\\
\end{tabular}}
\end{center}
\end{table*}
%

\textbf{PASCAL VOC 2007 Results.}
The results of the PASCAL VOC 2007 dataset evaluation appear in
Table~\ref{t:voc2007}. For the What Net, we utilized VGG16
pretrained on the ImageNet dataset~\cite{imagenet}.
We removed the fully connected
layers and softmax calculation from VGG16, and we calculated GT based
on the last convolutional layer. No fine tuning of parameters was
done. For the Where Net we used Resnet 101~\cite{resNet}. We also tried YOLO V2 for the Where Net. All networks were trained using 4 GPUs. 

We compared our performance results with those reported for a variety
of state-of-the-art approaches to object detection. Our primary
baseline was Faster-RCNN using a Resnet 101 network trained on PASCAL
VOC 2007. As shown in Table~\ref{t:voc2007}, the selective attention
process of our approach (What-Where Nets) resulted in substantially better performance in comparison to Faster-RCNN and other methods.  WW-Nets appear to be more accurate at detecting larger objects than smaller ones, perhaps because the region proposal network based its output on
the last (lowest resolution) convolutional layer. For most of the
object classes,  WW-Nets performed favorably against other methods by a large margin. One shot detectors did not perform as well as the region
proposal based approaches (as expected). We observed this by training YOLO V2 on PASCAL VOC 2007 and the union of PASCAL VOC 2007 and 2012 dataset and using the trained YOLO V2 network for the Where Net. The
difference likely reflects a trade off between speed and accuracy. \vspace{2mm}

\begin{table*}[t]
\caption{Comparison on COCO test-dev
reveals that WW-Net performs favorably against state-of-the-art methods. \label{t:coco}} 
\begin{center}
\scalebox{0.99}{
 \begin{tabular}{c|c|cccc} 
 \hline
Method & Backbone& Input Resolution & AP&$\text{AP}_{50}$&$\text{AP}_{75}$ \\
\hline
Faster R-CNN w/ FPN \cite{24_coco} &ResNet-101 &$1000 \times 600$&49.5& 59.1& 39.0 \\
Deformable-CNN \cite{7_coco}& Inception-ResNet &$1000 \times 600$&-& 58.0 & - \\
Deep Regionlets \cite{51_coco} &ResNet-101 &$1000 \times 600 $&-& 59.8 & - \\
YOLOv2 \cite{yolo9000} &DarkNet-19&$544 \times 544$&31.6& 44.0& 19.2 \\
YOLOv3 \cite{yolov3_coco}& DarkNet-53& $608 \times 608$&46.1& 57.9& 34.4\\
SSD \cite{ssd} &ResNet-101& $513 \times 513$&41.8&50.4& 33.3 \\
DSSD \cite{dssd} &ResNet-101& $513 \times 513$ &44.2& 53.3& 35.2 \\
RetinaNet \cite{retina} &ResNet-101& $1333 \times 800$&50.7&59.1 & 42.3 \\
WW-Net (ours) &Resnet 101 & $511 \times 511$&51.6& 57.8 &45.3 \\
\end{tabular}}
\end{center}
\end{table*}

\textbf{PASCAL VOC 2012 Results.}
We also measured WW-Nets performance on the PASCAL VOC 2012
dataset. The What Net was VGG16, with features for calculating
GT extracted from the last convolutional layer. The Where Net was
initialized with parameters previously learned for the PASCAL VOC 2007
evaluation, but further training was done on PASCAL VOC trainval sets. For the additional
training, the learning rate was initialized to $3 \times 10^{-4}$ for
$900,000$ iterations, and then it was reduced to $3 \times 10^{-5}$ until
reaching iteration $1,200,000$. The learning rate was further reduced to
$3 \times 10^{-6}$ until reaching $3,000,000$ iteration. The 
training was done on 4 GPUs. This resulted in
WW-Nets that produced comparable or better performance than
state-of-the-art methods. Performance results for PASCAL VOC 2012 are
shown in Table~\ref{t:voc2012}.

\vspace{2mm}
\textbf{COCO Results.}
We also measured WW-Nets performance on the COCO 
dataset. The What Net was VGG16, with features for calculating
GT extracted from the last convolutional layer. The Where Net was Resnet 101.
Initialized with parameters previously learned for the PASCAL VOC 2012
evaluation, the network was further trained using the COCO training-dev set. The 
training was done on 4 GPUs. This resulted in
WW-Nets that produced comparable or better performance than
state-of-the-art methods, often by a large margin. Performance results for COCO are
shown in Table~\ref{t:coco}.
\section{Human Attention}
\textbf{Participants.} $15$ healthy undergraduate students
($12$ female, $3$ male; age: mean$\pm$s.d.$=20.06\pm1.62$) were
recruited from the University of California, Merced campus.
Participants provided informed consent in accordance with IRB
protocols and received one hour of course credit for their
participation. Participation was restricted to those who reported
normal, uncorrected vision in a pre-screen survey.

\begin{figure}[hbt]
  \centering
  \begin{tabular}{c@{\hspace{2ex}}c@{\hspace{2ex}}}
    \rotatebox{90}{\hspace{19ex}MAE} &
    \includegraphics[width=0.85\linewidth]{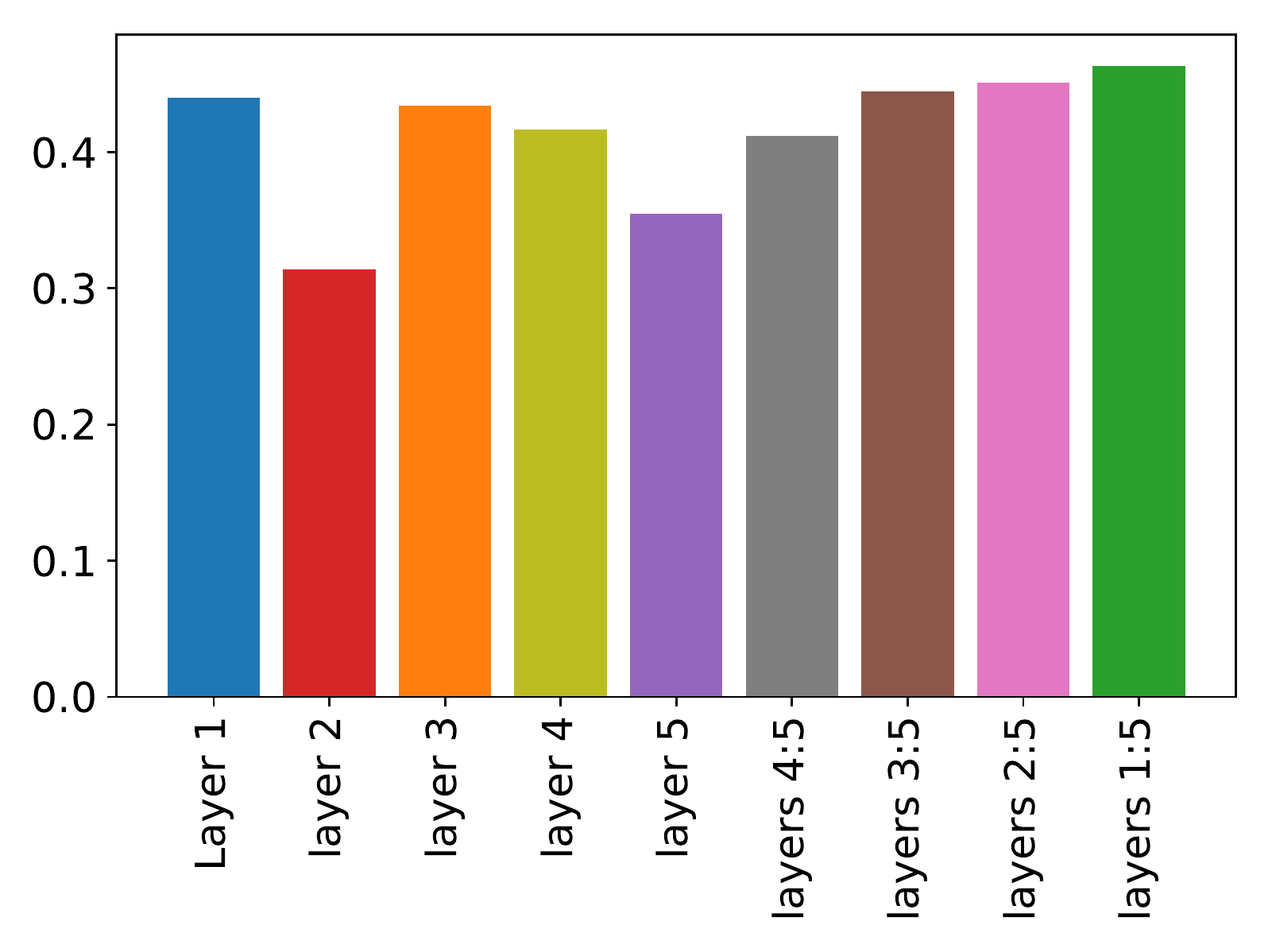}
  \end{tabular}
  \caption{MAE Measure Between Human Fixation Distributions and
           Attention Maps at Different Layers of WW-Nets}
  \label{f:perf}
\end{figure} 

\textbf{Materials.} The images seen were a subset of $200$ images from
PASCAL VOC 2007. For display, images were scaled up to double their
original size. The display width of images ranged from $636-1000$
pixels (mean$= 953.68$; median$=1000$; standard deviation=102.54). The
display height of images ranged from $350-1000$ pixels (mean$=759.71$;
median$=750$; standard deviation$=120.73$). The ordering of images was
randomized within participants. 

\textbf{Procedure.} Participants completed the task individually in a
research laboratory. Participants were seated at a desk in front of a
computer and were fitted with a head-mounted Eyelink II eye tracker
system. A microphone was placed nearby to record participants'
speech. Prior to the beginning of the experiment, the eye-tracker was
calibrated using a standard nine-point grid, and the subject was shown
how to perform a drift correction, which took place at the beginning
of each trial. Eye movement data was collected via the Eyelink control
software and custom MATLAB scripts. Data from the right eye were
collected using both pupil shape and corneal reflection.
Each trial began with the participant fixating the center of the
screen and pressing the space bar to initiate the trial. Then, an
image was displayed in the center of the screen for 5
seconds. Participants were instructed to name out loud as many
different objects as they could identify in the image within the
5-second time limit.\vspace{2mm}

\textbf{Fixation Heat-Map.} The raw eye-tracking data were converted
into MatLab data structures using the \texttt{Edf2Mat} package. Heat
maps were generated from the eye-fixation data. Fixations were
included in the analysis only if they began at or after the start of
the trial. Fixations were pooled from all participants. For each
image, a zero matrix with the same $N\times M$ dimensions as the
original image pixel height and width was created. Fixation
coordinates were scaled so that they corresponded to locations within
the original image sizes and then rounded to the nearest
integer. Thus, the coordinate values for each fixation corresponded 
to a location within the matrix. For each fixation, the value of the
corresponding matrix position was increased by the duration of the
fixation in milliseconds. Then, all values of the matrix were divided
by the maximum value of the matrix in order to normalize all matrix
values to the $[0,1]$ range. Finally, a convolution was performed on
the matrix using a Gaussian kernel ($\sigma = 20$, size = $80 \times
80$). These steps yielded a heat map showing the likely places to
which participants attend within the images. Calculations were
performed in MATLAB using custom scripts.\vspace{2mm}

\textbf{Eye-Tracking Study Results.}
Example distribution heatmaps of human fixations are shown in
Figure~\ref{f:HA}. We used these distributions as a form of
ground-truth for analyzing the attention maps produced by 
 WW-Nets. We compared the attention map distributions, $G(x,y)$, with the
human fixation distributions, $S(x,y)$, using a simple Mean Absolute Error (MAE) measure.
\begin{equation}
MAE = \frac{1}{W \times H} \sum_{x=1}^W \sum_{y=1}^H |S(x,y)-G(x,y)|
\end{equation}
where $W$ and $H$ are the width and height of the image. We performed
this assessment for various attention maps in the network. The resulting error values are shown in Figure~\ref{f:perf}.

We found that the second convolutional layer displayed the greatest
similarity to human performance. This contrasts with the attention
mechanisms in other object detection frameworks, which frequently base
attention only on the last convolutional layer. It appears as if the
high resolution and fundamental features learned by the network
provide relatively good guides for attention during object detection.
Both the object detection system software and
the fixation distribution data collected from human subjects will be
made publicly available. 

\section{Conclusion}
\label{s:conclusion}

In this paper, we highlighted the utility of incorporating 
selective attention mechanisms into object detection algorithms. We
suggested that such mechanisms could guide the search over image
regions, focusing this search in an informed manner. In addition, we
demonstrated that the resulting removal of distracting irrelevant
material can improve object detection accuracy substantially. Our approach was inspired by the visual system of the human
brain. Theories of spatial attention that see it as arising from dual
interacting ``what'' and ``where'' visual streams led us to propose a
dual network architecture for object detection. The resulting
architecture, WW-Nets, integrates attention based object detection methods with supervised approaches.

The benefits of selective attention, as implemented in WW-Nets, are evident in the performance results reported on the PASCAL VOC 2007, 
PASCAL VOC 2012, and COCO datasets. Evaluation experiments revealed that WW-Nets display greater object detection accuracy than state-of-the-art
approaches, often by a large margin.

\bibliographystyle{IEEEtran}
\bibliography{biblography}
\end{document}